\documentclass[10pt, a4paper]{article}

\usepackage{lrec}
\usepackage{multibib}
\newcites{languageresource}{Language Resources}
\usepackage{soul}

\usepackage{epstopdf}
\usepackage[latin1]{inputenc}

\usepackage{amsmath}
\usepackage[]{hyperref}
\usepackage{lipsum}
\usepackage{xspace}
\usepackage{xstring}
\usepackage{booktabs}
\usepackage{url}
\usepackage{microtype}
\usepackage{color}
\usepackage{relsize}
\usepackage{multirow}
\usepackage{multicol}
\usepackage{enumitem}
\usepackage{pifont}
\usepackage{tabu}
\usepackage{float}
\usepackage[outline]{contour}

\usepackage{aplcomments}

\usepackage{tikz}

\usepackage{graphicx}
\usepackage{tabularx}

\newcommand{\rom}[1]{%
	\textup{\uppercase\expandafter{\romannumeral#1}}%
}

\newcommand{\highest}[1]{\textbf{#1}}
\newcommand{\topk}[1]{top-$#1$}
\newcommand{\Tacc}[1]{$\text{Acc}_{\text{T}}^{#1}$\xspace}
\newcommand{\Facc}[1]{$\text{Acc}_{\text{F}}^{#1}$\xspace}
\newcommand{\Tiaa}{$\alpha_{\text{T}}$\xspace}
\newcommand{\Fiaa}{$\alpha_{\text{F}}$\xspace}
\newcommand{\TM}[1]{$\text{TM}^{#1}$\xspace}
\newcommand{\BLEU}[1]{$\text{BLEU}^{#1}$\xspace}

\newcommand{\eat}[1]{\ignorespaces}

\newcommand{\red}[1]{\textcolor{red}{#1}}

\newcommand{\wrong}[1]{\red{\underline{#1}}}
\newcommand{\incomplete}{\red{$\hdots$ \textit{incomplete}}}

\newcolumntype{L}[1]{>{\raggedright\let\newline\\\arraybackslash\hspace{0pt}}m{#1}}
\newcolumntype{C}[1]{>{\centering\let\newline\\\arraybackslash\hspace{0pt}}m{#1}}
\newcolumntype{R}[1]{>{\raggedleft\let\newline\\\arraybackslash\hspace{0pt}}m{#1}}

\def\|#1|{\mathid{#1}}
\newcommand{\mathid}[1]{\ensuremath{\mathit{#1}}}
\def\<#1>{\codeid{#1}}
\protected\def\codeid#1{\ifmmode{\mbox{\smaller\ttfamily{#1}}}\else{\smaller\ttfamily
		#1}\fi}

\newcommenter{chenglong}{1.0,0.4,1.0} 
\newcommenter{mde}{0.4,0.4,1.0} 
\newcommenter{vic}{1.0,0.4,0.4} 
\newcommenter{luke}{0.4,1,0.4} 
\newcommenter{deric}{1.0,1.0,0.4} 

\newcommand{\todo}[1]{{\color{red}\bfseries [[#1]]}}

\newcommand{\specialcell}[2][l]{%
	\begin{tabular}[#1]{@{}l@{}}#2\end{tabular}}

\newcommand{\ttlcb}{\texttt{\char "7B}}
\newcommand{\ttrcb}{\texttt{\char "7D}}
\newcommand{\ttcbs}{\ttlcb\ttrcb}

\usepackage{cleveref} 
\crefformat{chapter}{\S#2#1#3}
\crefmultiformat{chapter}{\S\S#2#1#3}{and~#2#1#3}{, #2#1#3}{, and~#2#1#3}
\crefformat{section}{\S#2#1#3}
\crefmultiformat{section}{\S\S#2#1#3}{and~#2#1#3}{, #2#1#3}{, and~#2#1#3}
\Crefformat{chapter}{Chapter~#2#1#3}
\Crefmultiformat{chapter}{Chapters~#2#1#3}{and~#2#1#3}{, #2#1#3}{, and~#2#1#3}
\Crefformat{section}{Section~#2#1#3}
\Crefmultiformat{section}{Sections~#2#1#3}{and~#2#1#3}{, #2#1#3}{, and~#2#1#3}
\crefname{algocf}{alg.}{algs.}
\Crefname{algocf}{Algorithm}{Algorithms}

\title{NL2Bash: A Corpus and Semantic Parser for \\
Natural Language Interface to the Linux Operating System}

\name{Xi Victoria Lin*\thanks{* Work done at the University of 
Washington.}, 
Chenglong Wang, Luke Zettlemoyer, Michael D. Ernst}

\address{Salesforce Research, University of Washington, University of 
Washington, University of Washington\\
xilin@salesforce.com, \{clwang,lsz,mernst\}@cs.washington.edu\\}

\abstract{We present new data and semantic parsing methods for the problem of mapping English sentences to Bash commands (NL2Bash). Our long-term goal is to enable any user to perform operations such as file manipulation, search, and application-specific scripting by simply stating their goals in English. We take a first step in this domain, by providing a new dataset of challenging but commonly used Bash commands and expert-written English descriptions, along with baseline methods to establish performance levels on this task.
 \\ \newline \Keywords{Natural Language Programming, Natural 
 Language Interface, Semantic Parsing} }

\begin{document}

\maketitleabstract

\section{Introduction}
\label{sec:intro}

The dream of using English or any other natural language to program computers 
has existed for almost as long as the task of programming 
itself~\cite{sammet1966use}. 
Although significantly less precise than a formal 
language~\cite{DBLP:conf/pc/Dijkstra78f}, natural 
language as a programming medium would be universally accessible and would 
support the automation of highly repetitive tasks such as file manipulation, 
search, and application-specific
scripting~\cite{Wilensky:1984:TUE:358080.358101,Wilensky:1988:BUC:65120.65123,dahl:1994:esa:1075812.1075823,dblp:conf/acl/quirkmg15,desai:2016:psu:2884781.2884786}.

This work presents new data and semantic parsing methods on a 
novel and ambitious domain --- natural language control of the operating system.
Our long-term goal is to enable any user to perform tasks on their
computers by simply stating their goals in natural language (NL). We take a 
first step in
this direction, by providing a large new dataset (NL2Bash) of challenging
but commonly used commands and expert-written descriptions, along with baseline 
methods
to establish performance levels on this task.  

The NL2Bash problem can be seen as a type of semantic parsing, where the goal 
is to map sentences to formal representations of their underlying 
meaning~\cite{mooney2014semantic}. 
The dataset we introduce provides a new 
type of target meaning representations (Bash\footnote{The Bourne-again 
	shell (Bash) is the most popular Unix shell and command language: 
	\url{https://www.gnu.org/software/bash/}. Our data collection approach and 
	baseline models can be trivially generalized to other command languages.} 
	commands), 
and is significantly larger (from two to ten times) than most existing semantic 
parsing 
benchmarks~\cite{dahl:1994:esa:1075812.1075823,popescu:2003:ttn:604045.604070}.
Other recent work in semantic parsing has also focused on 
programming languages, including regular 
expressions~\cite{DBLP:conf/emnlp/LocascioNDKB16}, IFTTT 
scripts~\cite{dblp:conf/acl/quirkmg15}, and SQL 
queries~\cite{kwiatkowski-etal:2013:emnlp,DBLP:conf/acl/IyerKCKZ17,zhong2017seq2sql}.
However, the shell command data we consider raises unique challenges, due to 
its irregular syntax (no syntax tree 
representation for the command options), wide domain coverage ($>$ 100 Bash 
utilities), and a large percentage of unseen words (e.g. commands can 
manipulate arbitrary files). 

We constructed the NL2Bash corpus with frequently used Bash commands scraped 
from websites such as question-answering forums, tutorials, tech blogs, and 
course materials. 
We gathered a set of high-quality 
descriptions of the commands from Bash programmers. Table~\ref{tb:examples} 
shows several examples. 
After careful quality control, we were able to gather over 9,000 
English-command pairs, covering over 100 unique
Bash utilities. 

We also present a set of experiments to demonstrate that NL2Bash is a 
challenging task which is worthy of future study. 
We build on recent work in neural semantic 
parsing~\cite{dong-lapata:2016:p16-1,DBLP:conf/acl/LingBGHKWS16}, by evaluating 
the standard 
Seq2seq model~\cite{Sutskever:2014:SSL:2969033.2969173} and the 
CopyNet model~\cite{DBLP:conf/acl/GuLLL16}.
We also include a recently proposed stage-wise neural semantic parsing model, 
Tellina, which uses manually defined heuristics for 
better detecting and translating the command arguments~\cite{tellina}.
We found that when applied at the right sequence granularity (sub-tokens), 
CopyNet 
significantly outperforms the stage-wise model, with significantly less 
pre-processing and 
post-processing. Our best performing system obtains top-1 
command structure accuracy of 49\%, and top-1 full command accuracy of 36\%. 
These performance levels, although far from perfect, are high enough to be 
practically useful
in a well-designed interface~\cite{tellina}, and also suggest ample room for 
future modeling innovations. 
\section{Domain: Linux Shell Commands}
\label{sec:domain}

\begin{table*}[ht]
	\begin{center}
		\begin{tabular}{m{5cm}|l}
			\hline
			Natural Language & Bash Command(s)\\
			\hline
			\emph{find .java files in the current directory tree that contain
				the pattern `TODO' and print their names} & \specialcell{
				\<grep -l "TODO" *.java> \\
				\<find .\ -name "*.java" -exec grep -il "TODO" \ttcbs\
				$\backslash$;> \\
				\<find .\ -name "*.java" | xargs -I \ttcbs\ grep -l "TODO" 
				\ttcbs>
			} \\
			\hline
			\emph{display the 5 largest files in the current directory and its
				sub-directories} & \specialcell{
				\<find . -type f | sort -nk 5,5 | tail -5> \\
				\<du -a .\ | sort -rh | head -n5> \\
				\<find .\ -type f -printf '$\%$s $\%$p$\backslash$n' | sort -rn
				| head -n5> 
			} \\
			\hline
			\emph{search for all jpg images on the system and archive them to
				tar ball ``images.tar''} & \specialcell{
				\<tar -cvf images.tar \$(find / -type f -name *.jpg)> \\
				\<tar -rvf images.tar \$(find / -type f -name *.jpg)> \\
				\<find / -type f -name "*.jpg" -exec tar -cvf images.tar
				\ttcbs\ $\backslash$;> \\
			} \\
			\hline
		\end{tabular}
	\end{center}
	\caption{Example natural language descriptions and the corresponding
		shell commands from NL2Bash.} 
	\label{tb:examples}
\end{table*}

A shell command consists of three basic components, as shown in 
Table~\ref{tb:examples}: 
utility (e.g. \<find>, \<grep>), option flags (e.g. \<-name>, 
\<-i>),
and arguments (e.g. \<"*.java">, \<"TODO">).  
A utility can have 
idiomatic syntax for flags (see the \<-exec $\hdots$ \ttcbs\ $\backslash$;> 
option of
the \<find> command).


There are over 250 Bash utilities, and new ones are regularly added by 
third party developers. We focus on 135 of the most useful utilities identified 
by the Linux user group 
(\url{http://www.oliverelliott.org/article/computing/ref_unix/}), that is, our 
domain of target commands contain only those 135 utilities.\footnote{We were 
	able to gather fewer examples for the less common ones. Providing the 
	descriptions for them also requires a higher level of Bash expertise of the 
	corpus annotators.}
We only considered the target commands that can be 
specified in a single line (one-liners).\footnote{We decided 
	to investigate this simpler case prior to synthesizing longer
	shell scripts because one-liner Bash commands are practically useful and 
	have 
	simpler structure. Our baseline results and analysis (\cref{sec:baselines}) 
	show that even this task is challenging.} 
Among them, we omitted commands that contain syntax structures such as I/O 
redirection, variable assignment, and compound statements because those 
commands 
need to be interpreted in context.
Table~\ref{tb:scope} summarizes the in-scope and out-of-scope syntactic 
structures of the shell commands we considered. 

\begin{table}
	\begin{center}
		\begin{tabular}{m{1.3cm}|m{6cm}}
			In-scope & \specialcell{
				1. Single command \\
				2. Logical connectives: \<\&\&>, \<||>, parentheses \<()> \\
				3. Nested commands: \\
				\ \ \ \ - pipeline \<|>	\\
				\ \ \ \ - command substitution \<\$()>	\\
				\ \ \ \ - process substitution \codeid{<()}	
			}\\
			\hline
			Out-of-scope & \specialcell{
				1. I/O redirection \codeid{<}, \codeid{<<} \\
				2. Variable assignment \<=> \\
				3. Compound statements: \\
				\ \ \ \ - \<if>, \<for>, \<while>, \<util> statements \\
				\ \ \ \ - functions \\
				4. Non-bash program strings nested with \\
				language interpreters such as \<awk>, \<sed>, \\
				\<python>, \<java>
			}
		\end{tabular}
		\caption{In-scope and out-of scope syntax for the Bash commands in our 
		dataset.}
		\label{tb:scope}
	\end{center}
\end{table}
\section{Corpus Construction}
\label{sec:data}

The corpus consists of 
text--command pairs, where each pair consists of a 
Bash command scraped from the web and an expert-generated natural language description.
Our dataset is publicly available for use by other researchers: 
\url{https://github.com/TellinaTool/nl2bash/tree/master/data}.

We collected 12,609 text--command pairs in total (\cref{subsec:data-collection}).
After filtering, 9,305 pairs remained (\cref{subsec:data-filter}).
We split this data into train, development (dev), and test sets, subject to the constraint that neither a natural language description nor a Bash command appears in more than one split (\cref{subsec:data-split}).

\subsection{Data Collection}
\label{subsec:data-collection}

We hired 10 Upwork\footnote{\url{http://www.upwork.com/}} freelancers who are familiar with shell scripting.
They collected text--command pairs from web pages such as question-answering forums, tutorials, tech blogs, and course materials. 
We provided them a web inferface to assist with searching, page browsing, and
data entry.

The freelancers copied the Bash command from the webpage, and either copied the text from the webpage or wrote the text based on their background
knowledge and the webpage context.
We restricted the natural language description to be a single 
sentence and the Bash command to be a one-liner. We found that oftentimes one 
sentence is enough to accurately describe the function of the 
command.\footnote{As discussed in \cref{subsec:error}, in 4 out of 
100 examples, a one-sentence description is difficult to interpret. 
Future work should
investigate interactive natural language programming approaches in these 
scenarios.}

The freelancers provided one natural-language description for each command on a
webpage.  A freelancer might annotate the same command multiple times in
different webpages, and multiple freelancers might annotate the same
command (on the same or different webpages).
Collecting multiple different descriptions increases language diversity in the
dataset. On average, each freelancer collected 50 pairs/hour.

\subsection{Data Cleaning}
We used an automated process to filter and clean the dataset, as described
below.  Our released corpus includes the filtered data, the full
  data, and the cleaning scripts.

\paragraph{Filtering}
\label{subsec:data-filter}
The cleaning scripts removed the following commands.
\begin{itemize}[topsep=0pt,itemsep=-1ex,partopsep=1ex,parsep=1ex]
\item Non-grammatical commands that violate the syntax specification in
the Linux man pages (\url{https://linux.die.net/man/}).
\item Commands that contain out-of-scope syntactic structures shown in 
\Cref{tb:scope}.
\item Commands that are mostly used in multi-statement
 shell scripts (e.g. \<alias> and \<set>).
\item Commands that contain non-bash language interpreters (e.g.
\<python>, \<c++>, \<brew>, \<emacs>).
These commands contain strings in other programming languages.
\end{itemize}

\paragraph{Cleaning} 
We corrected
spelling errors in the natural language descriptions using a probabilistic spell
checker (\url{http://norvig.com/spell-correct.html}).
We also manually corrected a subset of the spelling errors that bypassed the
spell checker in both the natural language and the shell commands.
For Bash commands, we removed \<sudo> and the shell input prompt characters such as
``\$'' and ``\#'' from the beginning of each command.  We replaced the
absolute pathnames for utilities by their base names (e.g., we changed \</bin/find> 
to
\<find>). 

\subsection{Corpus Statistics}
\label{subsec:data-stats}

After filtering and cleaning, our dataset contains 9,305 pairs.
The Bash commands cover 102 unique utilities using 206 flags
--- a rich functional domain.

\paragraph{Monolingual Statistics} Tables~\ref{tbl:nl-stats} 
and~\ref{tbl:bash-stats} show the statistics of natural language (NL) and Bash 
commands in our corpus.  
	
\begin{figure}[ht]
	\centering
	\includegraphics[width=0.97\linewidth]{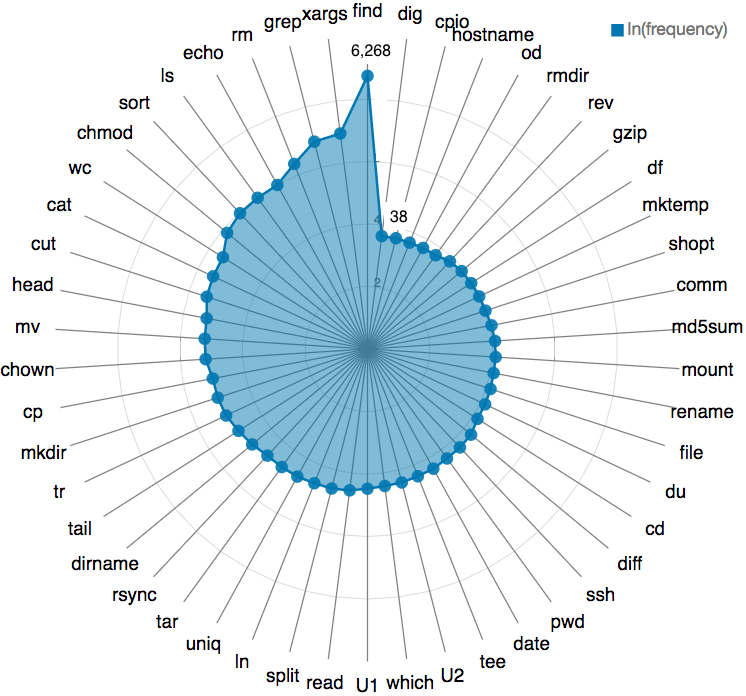}
	\caption{Top 50 most frequent bash utilities in the dataset with
	  their frequencies in log scale. \<U1> and \<U2> at the bottom of the 
	  circle denote the utilities \<basename> and \<readlink>.
	}
	\label{fig:utility-freq}

\eat{
	\todo{Provide a vector-graphics version of this diagram.  It's currently a
  	bitmap, which is very hard to read.}
	\todo{Change the numbers 2,4,6,8,10 to actual numbers.  That is, even if
  	the graph is log-scale, the axis numbers should be actual measurments,
  	not logarithms.}
	\todo{It's weird that this diagram is circular but 
	\cref{fig:utility-flag-number} represents the same data as a histogram.  I
	would make the two consistent (use a histogram here).}
	\todo{Ensure that the actual number can be determined from this diagram,
	just as it can for \cref{fig:utility-flag-number}. (The units aren't even 
	labeled on this.)}
}
\end{figure}

The average length of the NL sentences and Bash commands are 
relatively short, being 11.7 words and 7.7 tokens respectively. The 
median word frequency and command token frequency are both 1, 
which is caused by the large number of open-vocabulary constants (file names, 
date/time expressions, etc.) that appeared only once in the 
corpus.\footnote{As shown in figure~\ref{fig:utility-freq}, the 
most frequent bash utilities appeared over 6,000 times in the corpus. 
Similarly, natural language words such as ``files'', ``in'' appeared in 
5,871 and 5,430 sentences, respectively. These extremely high frequency tokens 
are the reason for the significant difference between the averages and
medians in Tables~\ref{tbl:nl-stats} and~\ref{tbl:bash-stats}.} 

We define a command template as a command with its arguments 
	replaced by their semantic types. For example, the template of \<grep -l 
	"TODO" *.java> is \<grep -l [regex] [file]>.

\begin{table}[t]
	\centering
	\begin{tabular}{c|c|c|c|c|c}
		\hline
		\multirow{2}{*}{\# sent.} & \multirow{2}{*}{\# word} & 
		\multicolumn{2}{c|}{\# words per sent.} & 
		\multicolumn{2}{c}{\# sent. per word} \\
		\cline{3-4} \cline {5-6}
		& & avg. & median & avg. & median \\
		\hline
		8,559 & 7,790 & 11.7 & 11 & 14.0 & 1 \\
		\hline
	\end{tabular}
	\caption{Natural Language Statistics: \small{\# unique sentences, \# unique 
	words, 
	\# words per sentence and \# sentences that a word appears 
	in}.}\label{tbl:nl-stats}
\end{table}

\begin{table}
	\centering
	\setlength\tabcolsep{3pt}
	\begin{tabular}{c|c|c|c|c|c|c}
		\hline
		\multirow{2}{*}{\# cmd} & \multirow{2}{*}{\# temp} & 
		\multirow{2}{*}{\# token} & 
		\multicolumn{2}{c|}{\# tokens / cmd} & 
		\multicolumn{2}{c}{\# cmds / token} \\
		\cline{4-5} \cline {6-7}
		& & & avg. & median & avg. & median \\
		\hline
		7,587 & 4,602 & 6,234 & 7.7 & 7 & 11.5 & 1 \\
		\hline
	\end{tabular}
	\newline
	\vspace*{.3cm}
	\newline
	\begin{tabular}{c|c|c|c|c|c|c}
		\hline
		\multirow{2}{*}{\# utility} & \multirow{2}{*}{\# flag} & 
		\# reserv. & \multicolumn{2}{c|}{\# cmds / util.} 
		& \multicolumn{2}{c}{\# cmds / flag} \\
		\cline{4-5} \cline {6-7}
		& & token & avg. & median & avg. & median \\
		\hline
		102 & 206 & 15 & 155.0 & 38 & 101.7 & 7.5  \\
		\hline
	\end{tabular}
	\caption{Bash Command Statistics.  {\small The top table contains \# unique 
	commands, \# unique command templates, \# unique 
	tokens, \# tokens per command and \# commands that a token appears in. The 
	bottom table contains \# unique utilities, \# unique flags, \# unique 
	reserved tokens, \# commands a utility appears in and \# commands a flag 
	appears in}.}.\label{tbl:bash-stats}
\end{table}

\paragraph{Mapping Statistics}
Table~\ref{tb:nl2bash-stats} shows the statistics of natural language to Bash command mappings in our dataset. While most of the NL sentences and Bash commands form one-to-one mappings, the problem is naturally a many-to-many 
mapping problem --- there exist many semantically equivalent commands, and one Bash command may be phrased in different NL descriptions. 
This many-to-many mapping is common in machine translation 
datasets~\cite{papineni:2002:bma:1073083.1073135}, but rare for traditional 
semantic parsing 
ones~\cite{dahl:1994:esa:1075812.1075823,Zettlemoyer:2005:LMS:3020336.3020416}.

As discussed in \cref{sec:eval} and \cref{subsec:results}, the 
many-to-many mapping affects both evaluation and modeling choices.

\begin{table}[t]
	\centering
	\begin{tabular}{c|c|c|c|c|c}
		\hline
		\multicolumn{3}{c|}{\# cmd per nl} & \multicolumn{3}{c}{\# nl
			per cmd} \\
		\cline{1-3}\cline{4-6}
		avg. & median & max & avg. & median & max \\
		\hline
		1.09 & 1 & 9 & 1.23 & 1 & 22 \\
		\hline
	\end{tabular}
	\caption{Natural Language to Bash Mapping 
		Statistics}\label{tb:nl2bash-stats}
\end{table}

\paragraph{Utility Distribution}  
Figure~\ref{fig:utility-freq} shows the top 50 most common Bash utilities
in our dataset and their frequencies in log-scale.
The distribution is long-tailed: the top most frequent utility \<find> appeared 
6,268 times and the second most frequent utility \<xargs> appeared 1,047 times. 
The 52 least common bash utilities, in total, appeared only 984 times.\footnote{
	The utility \<find> is disproportionately common in our corpus. This is 
	because we collected the data in two separated stages. As a proof 
	of concept, we initially collected 5,413 commands that 
	contain the utility \<find> (and may also
	contain other utilities). After that, we 
	allow the freelancers to collect all commands that contain any of the 135 
	utilities described in~\cref{sec:domain}.
}

The appendix (\S\ref{sec:appendix}) gives more corpus statistics.

\subsection{Data Split}
\label{subsec:data-split}
We split the filtered data into train, dev, and test sets
(Table~\ref{tb:data-split}).
We
first clustered the pairs by NL descriptions --- a cluster contains all pairs 
with the identical normalized NL description.
	We normalized an NL description by lower-casing, stemming, and stop-word 
	filtering, as described in~\cref{subsec:preprocessing}

We randomly split the clusters into train, dev, and test at a ratio of 10:1:1. 
After splitting, we moved all development and test pairs whose command 
appeared in the train set into the train set.
This prevents a model from obtaining high accuracy by
trivially memorizing a natural language description or a command it has seen
in the train set, which allows us to evaluate the model's ability to generalize.

\begin{table}[ht]
	\centering
	\begin{tabular}{c|c|c|c}
		\hline
		& Train & Dev & Test \\
		\hline
		\# pairs & 8,090 & 609 & 606 \\
		\# unique nls & 7,340 & 549 & 547 \\
		\hline
	\end{tabular}
	\caption{Data Split Statistics}\label{tb:data-split}
\end{table}


\section{Evaluation Methodology}
\label{sec:eval}

In our dataset, one natural language description may have multiple correct
Bash command translations.  This presents challenges for evaluation since not 
all 
correct commands are present in our dataset. 

\paragraph{Manual Evaluation}
We hired three Upwork freelancers 
who are familiar with shell scripting. 
To evaluate a particular system,
the freelancers independently evaluated the correctness of its \topk{3} 
translations for 
all test examples.
For each command translation, we use the majority vote of the three 
freelancers as the final evaluation. 

We grouped the test pairs that have the same normalized NL descriptions as a 
single test instance (Table~\ref{tb:data-split}). We report two types of 
accuracy: \topk{k} full command accuracy 
(\Facc{k}) and \topk{k} command template accuracy (\Tacc{k}).  
We define \Facc{k} to be the 
percentage of test instances for which a correct full command is ranked $k$ or 
above in the model output.
We define
\Tacc{k} to be the percentage of test instances for which a correct command 
template is ranked $k$ or above in the model output (i.e., ignoring
incorrect arguments).


Table~\ref{tb:iaa} shows the 
inter-annotator agreement between the three pairs of our freelancers on both the
template judgement (\Tiaa) and full-command judgement (\Fiaa).

\begin{table}[ht]
	\centering
	\begin{tabular}{c|c|c|c|c|c}
		\hline
		\multicolumn{2}{c|}{Pair 1} & \multicolumn{2}{c|}{Pair 2} & 
		\multicolumn{2}{c}{Pair 3} \\
		\cline{1-2}\cline{3-4}\cline{5-6} 
		\Fiaa & \Tiaa & \Fiaa & \Tiaa & \Fiaa & \Tiaa \\ 
		\hline
		0.89 & 0.81 & 0.83 & 0.82 & 0.90 & 0.89 \\
		\hline
	\end{tabular}
	\caption{Inter-annotator agreement.}\label{tb:iaa}
\end{table}

\paragraph{Previous approaches}
Previous NL-to-code translation work also noticed similar 
problems.  

\cite{dblp:conf/naacl/kushmanb13,DBLP:conf/emnlp/LocascioNDKB16} 
formally verify the equivalence of different regular expressions by 
converting them to minimal deterministic finite automaton (DFAs). 

Others~\cite{kwiatkowski-etal:2013:emnlp,DBLP:conf/acl/LongPL16,DBLP:conf/acl/GuuPLL17,DBLP:conf/acl/IyerKCKZ17,zhong2017seq2sql}
evaluate the generated code through execution.
As Bash is a Turing-complete language, verifying the equivalence of two
Bash commands is undecidable.
Alternatively, one can check command equivalence using test 
examples: two commands can be executed in a virtual environment and their 
execution outcome can be compared.
We leave this evaluation approach to 
future work.

Some other works~\cite{DBLP:conf/kbse/OdaFNHSTN15} have adopted fuzzy 
evaluation metrics, such as 
BLEU, which is widely used to measure the 
translation quality between natural 
languages~\cite{Doddington:2002:AEM:1289189.1289273}. 
Appendix~\ref{subsec:auto-eval} shows that the n-gram overlap 
captured by BLEU is not effective in measuring the semantic similarity for 
formal languages.


\section{System Design Challenges}
\label{sec:challenge}

%
%
This section lists challenges for semantic parsing in the Bash
domain.

\paragraph{Rich Domain} The application domain of Bash
ranges from file system management, text processing, network control
to advanced operating system functionality such as process management. Semantic 
parsing 
in Bash is equivalent to semantic parsing for each of the 
applications. In comparison, many previous works focus on only one 
domain (\cref{sec:comparison}).

\paragraph{Out-of-Vocabulary Constants} Bash commands contain many
open-vocabulary constants such as file/path names, file properties, time
expressions, etc. These form the unseen tokens for the trained
model. Nevertheless, a semantic parser on this domain should be able to
generate those constants in its output. This problem exists in nearly all
NL-to-code translation problems but is particularly severe for Bash 
(\cref{subsec:data-stats}).
What makes the problem worse is that oftentimes, the 
constants corresponding to the command arguments need to be properly 
reformatted following 
idiomatic syntax rules.

\paragraph{Language Flexibility} Many bash commands have a large set of option 
flags, and multiple commands can be 
combined to solve more complex tasks. 
This often results in multiple correct 
solutions for one task (\cref{subsec:data-stats}), and poses challenges for 
both training and evaluation.

\paragraph{Idiomatic Syntax} The Bash interpreter uses a shallow 
syntactic grammar to parse pipelines, code blocks, and other high-level syntax 
structures. 
It parses command options using pattern matching and 
each command can have idiomatic syntax rules (e.g. to 
specify an \<ssh> remote, the format needs to be
\<[USER@]HOST:SRC>). 
Syntax-tree-based parsing 
approaches~\cite{DBLP:conf/acl/YinN17,DBLP:conf/acl/GuuPLL17} 
are hence difficult to apply.
\section{Baseline System Performance}
\label{sec:baselines}
To establish performance levels for future work, we evaluated two neural 
machine translation models that have demonstrated strong performance in both 
NL-to-NL translation and NL-to-code translation tasks, namely, 
Seq2Seq~\cite{Sutskever:2014:SSL:2969033.2969173,dong-lapata:2016:p16-1}
 and CopyNet~\cite{DBLP:conf/acl/GuLLL16}. We also evaluated a stage-wise 
 natural language programing model, Tellina~\cite{tellina}, which includes 
 manually-designed heuristics for argument translation.
 
\label{subsec:sys-desp}
\paragraph{Seq2Seq} The Seq2Seq (sequence-to-sequence) model
defines the conditional probability of an output sequence given the input 
sequence using an RNN (recurrent neural network) encoder-decoder 
~\cite{Jain:1999:RNN:553011,Sutskever:2014:SSL:2969033.2969173}. 
When applied to the NL-to-code translation problem, the
input natural language and output commands are treated as sequences of 
tokens. At test time, the command sequences with the highest conditional 
probabilities 
were output as candidate translations.

\paragraph{CopyNet} CopyNet~\cite{DBLP:conf/acl/GuLLL16} is an extension of 
Seq2Seq which is able to select sub-sequences of the input sequence and emit 
them at proper places while generating the output sequence. 
The copy action is mixed with the regular token generation of the Seq2Seq 
decoder and the whole model is still trained end-to-end.

\paragraph{Tellina} The stage-wise natural language 
programing model, Tellina~\cite{tellina}, first abstracts the constants 
in an NL to their corresponding semantic types (e.g. \<File> and \<Size>) and performs template-level NL-to-code 
translation. It then fills the argument slots in the code template
with the extracted constants using a learned 
alignment model and reformatting heuristics. 

\subsection{Implementation Details}
\label{subsec:preprocessing}
We used the Seq2Seq formulation as specified in~\cite{Sutskever:2014:SSL:2969033.2969173}. 
We used the gated recurrent unit 
(GRU)~\cite{dblp:journals/corr/chunggcb14} RNN cells and a bidirectional 
RNN~\cite{schuster:1997:brn:2198065.2205129} encoder. We used the copying mechanism 
proposed by~\cite{DBLP:conf/acl/GuLLL16}.
The rest of the model architecture is the same as the Seq2Seq model.

We evaluated both Seq2Seq and CopyNet at three levels of token 
granularities: token, character and sub-token. 

\paragraph{Pre-processing} 
We used a simple regular-expression based natural language tokenizer and the 
Snowball stemmer to tokenize and stem the natural language. 
We converted all closed-vocabulary words in the natural language to lowercase 
and removed words in a stop-word list. We removed all NL 
tokens that appeared less than four times from the vocabulary for the token- and 
sub-token-based models. 
We used a Bash parser augmented from Bashlex 
(\url{https://github.com/idank/bashlex}) to parse and 
tokenize the bash commands. 

To compute the sub-tokens\footnote{As discussed in \cref{subsec:results}, 
	the simple sub-token based approach is surprisingly 
	effective for this problem. It avoids modeling 
	very long sequences, as the character-based models do, by preserving
	trivial compositionality in consecutive alphabetical letters and digits. On 
	the other hand, the separation between letters, digits, and special tokens 
	explicitly represented most of the idiomatic syntax of Bash we observed in 
	the data: the sub-token 
	based models effectively learn basic string manipulations (addition, 
	deletion and replacement of substrings) and the semantics of 
	Bash reserved tokens such 
	as \<\$>, \<">, \<*>, etc.}, we split every constant in both the 
	natural language and Bash commands into consecutive 
sequences of alphabetical letters and digits; all other characters are treated as an 
individual sub-token. (All Bash utilities and flags are treated as atomic tokens as they 
are not constants.) 
A sequence of sub-tokens as the result of a token split is padded 
with the special symbols \<SUB\_START> and \<SUB\_END> at the beginning and the end. For 
example, the 
file path ``/home/dir03/*.txt'' is converted to the sub-token sequence: 
\<SUB\_START>, ``/'', ``home'', ``/'', ``dir'', ``03'', ``/'', ``*'', ``.'', 
``txt'', \<SUB\_END>.

\paragraph{Hyperparameters}
The dimension of our decoder RNN is 400. The dimension of the two RNNs in 
the bi-directional encoder is 200. 
We optimized the learning objective with mini-batched
Adam~\cite{DBLP:journals/corr/KingmaB14}, using the default momentum 
hyperparameters. Our initial learning rate is
0.0001 and the mini-batch size is 128. We used variational RNN 
dropout~\cite{DBLP:conf/nips/GalG16} with 0.4 dropout rate. 
For decoding we set the beam size to 100.
The hyperparameters were set based on the model's performance on a development 
dataset (\cref{subsec:data-split}). 

Our baseline system implementation is released on Github: 
\url{https://github.com/TellinaTool/nl2bash}.

\subsection{Results}
\label{subsec:results}

\begin{table}[t]
	\centering
	\begin{tabular}{c|c|cc|cc}
		\hline
		\multicolumn{2}{c|}{Model} & \Facc{1} &\Facc{3}&\Tacc{1} &\Tacc{3}\\
		\hline
		\multirow{3}{*}{Seq2Seq} 
			& Char 	&0.24 	&0.27 	&0.35	&0.38 \\
			& Token	&0.10	&0.12 	&\highest{0.53}	&0.59\\
			& Sub-token &0.19	&0.27 	&0.41	&0.53	\\
		\hline
		\multirow{3}{*}{CopyNet} 
			& Char &0.25	&0.31	&0.34	&0.41	\\
			& Token &0.21	&0.34	&0.47	&\highest{0.61}	\\
			& Sub-token &\highest{0.31}	&\highest{0.40}	&0.44	&0.53 \\
		\hline
		\multicolumn{2}{c|}{Tellina} &0.29  &0.32	&0.51	&0.58\\
		\hline
	\end{tabular}
	\caption{Translation accuracies of the baseline systems on 100 instances 
	sampled from the dev set.}
	\label{tb:baselines}
\end{table}

Table~\ref{tb:baselines} shows the performance of the baseline
systems on 100 examples sampled from our dev set. Since manually evaluating all 
7 baselines on the complete dev set is expensive, we report the manual 
evaluation results on a sampled subset in Table~\ref{tb:baselines} and the 
automatic evaluation results on the full dev set in 
Appendix~\ref{subsec:auto-eval}.

Table~\ref{tb:sample-output} shows a few dev set examples and the baseline 
system translations. We now summarize the comparison between the different systems.

\paragraph{Token Granularity}
In general, token-level modeling yields higher command structure accuracy 
compared to using characters and sub-tokens. Modeling at the other two 
granularities gives higher full command accuracy.
This is expected since the character and sub-token models need to learn 
token-level compositions. They also operate over longer sequences 
which presents challenges for the neural networks.
It is somewhat surprising that Seq2Seq at the character level achieves 
competitive full command accuracy. However, the structure accuracy of these 
models is significantly lower than the other two 
counterparts.\footnote{~\cite{tellina} reported that incorrect commands 
can help human subjects, even when their arguments contain errors. This is 
because in many cases the human subjects were able to change or replace the 
wrong arguments based on their prior knowledge. 
Given this finding, we expect pure character-based models to be less useful in 
practice compared to the other two groups if we cannot find ways to improve 
their command structure accuracy.}

\paragraph{Copying}
Adding copying slightly improves the 
character-level models. This is expected as out-of-vocabulary characters are rare.
Using token-level copying improves full command accuracy significantly from 
vanilla Seq2Seq. However, the command template accuracy drops slightly, possibly due to the mismatch between the source constants and the command arguments, as 
a result of argument reformatting. We observe a
similarly significant full command accuracy improvement by adding copying at the 
sub-token level. The resulting ST-CopyNet model has the highest full 
command accuracy and competitive command template accuracy.

\paragraph{End-To-End vs. Pipline} 
The Tellina model which does template-level translation and argument 
filling/reformatting in a stage-wise manner yields the second-best full command 
accuracy and second-best structure accuracy. Nevertheless, the higher full 
command accuracy of ST-CopyNet (especially on the \Tacc{3} metrics) shows that 
learned string-level transformations out-perform manually written heuristics 
when enough data is provided. This shows the promise of applying 
end-to-end learning on such problems in future work.

Table~\ref{tb:baselines-test} shows the test set accuracies of the top-two 
performing approaches, ST-CopyNet and Tellina, evaluated on the entire test set. The accuracies of both models are higher than those on the dev 
set\footnote{One possible reason is that two 
	different sets of programmers evaluated the results on dev and test.}, 
but the relative performance gap holds:
ST-CopyNet performs significantly better than Tellina on the full 
command accuracy, with only a mild decrease in structure accuracy.	

\begin{table}[t]
	\centering
	\begin{tabular}{c|c|c|c|c}
		\hline
		Model & \Facc{1} &\Facc{3} &\Tacc{1} &\Tacc{3} \\
		\hline 
		ST-CopyNet & \highest{0.36} & \highest{0.45} & 0.49 & 0.61 \\
		\hline
		Tellina & 0.27 & 0.32 & 0.53 & 0.62 \\
		\hline
	\end{tabular}
	\caption{Translation accuracies of ST-CopyNet and Tellina on the full test 
		set.}
	\label{tb:baselines-test}
\end{table}	

Section~\ref{subsec:error} furthur discusses the comparison between these
two systems through error analysis.

\subsection{Error Analysis}
\label{subsec:error}
We manually examined the top-1 system outputs of ST-CopyNet and Tellina on the 
100 dev set examples and compared their error cases.

\begin{figure}[t]
	\centering
	\includegraphics[width=0.8\linewidth]{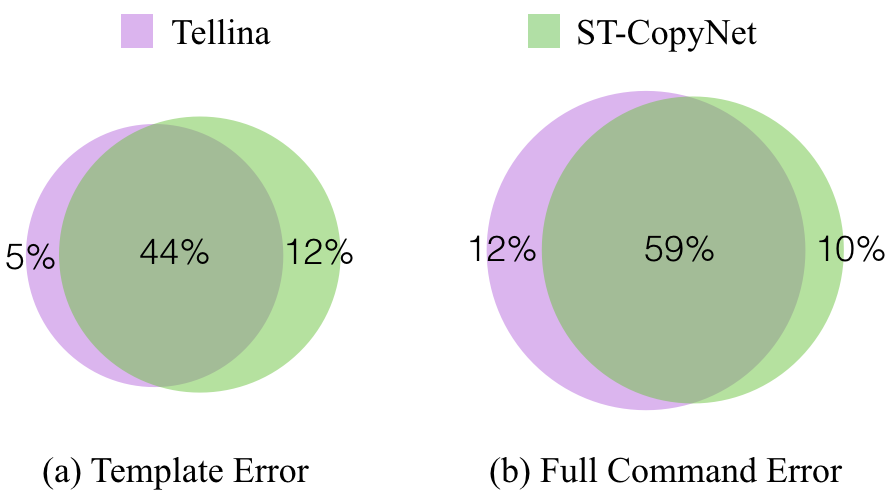}
	\caption{Error overlap of ST-CopyNet and 
		Tellina. The number denotes the percentage out of the 100 dev
		samples.}\label{fig:stcopy-tellina}
\end{figure}

\begin{figure}[t]
	\centering
	\includegraphics[width=\linewidth]{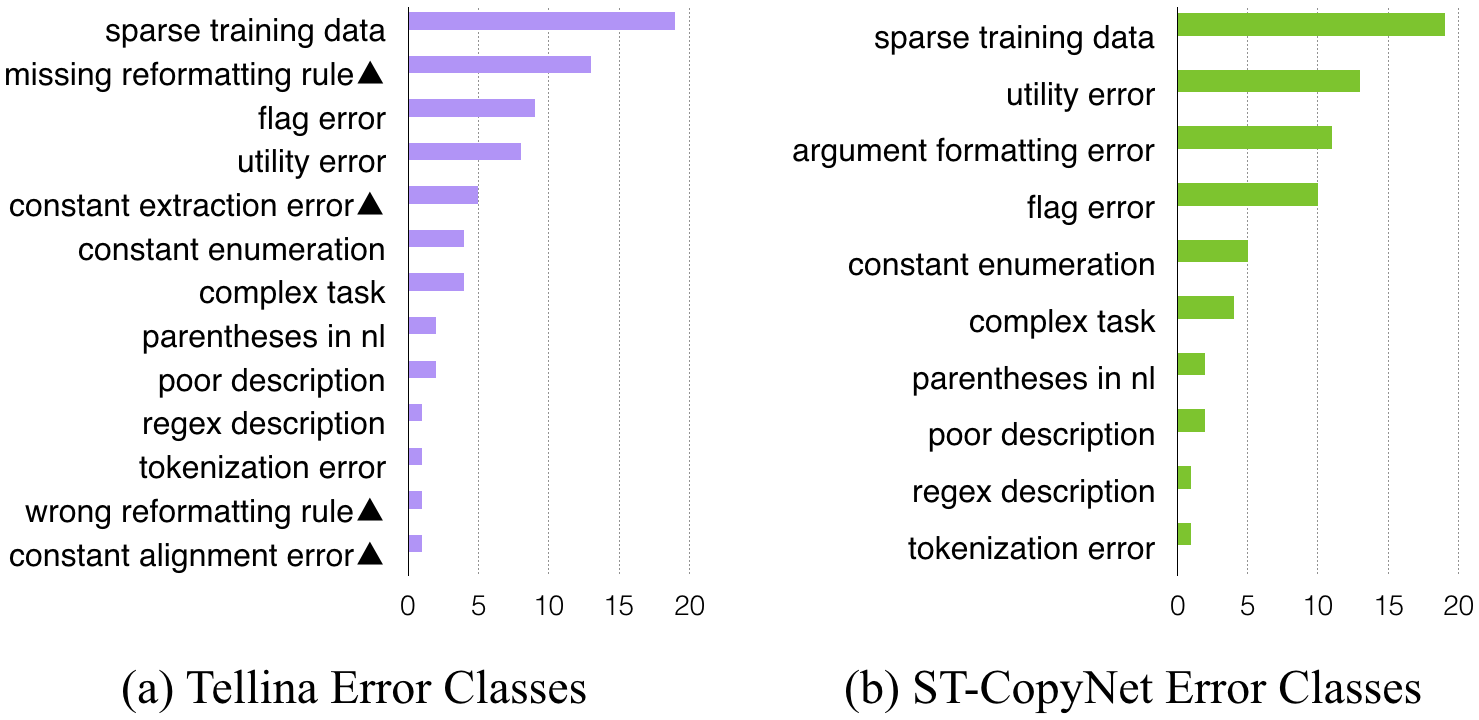}
	\caption{Number of error instances in each error classes of 
		ST-CopyNet and Tellina. The classes marked with \small{\ding{115}} 
		are unique to the pipeline system.}\label{fig:error-class}
\end{figure}

Figure~\ref{fig:stcopy-tellina} shows the error case overlap of the two
systems. For a significant proportion of the
examples both systems made mistakes in their translation (44\% by command structure error and 59\% by full 
command error). 
This is because the base model of the two systems are similar --- they are 
both RNN based models that perform sequential translation. Many such errors were caused by the NL 
describing a function that rarely appeared in the train set, or the GRUs failing to capture
certain portions of the NL descriptions.
For cases where only one of the models makes 
mistakes, Tellina makes fewer template errors and ST-CopyNet makes fewer full 
command errors.

We categorized the error causes of each system
(Figure~\ref{fig:error-class}), and discuss the major error classes below.

\begin{table}[t]
	\centering
	\begin{tabular}{l}
		\textbf{Sparsity in training data} \\
		\vspace{6pt}\hspace{.5cm} \specialcell{
			\textit{find all the text files in the file system 
				and \underline{search}} \\
			\textit{\underline{only in the disk partition of the root}}.} \\
		\textbf{Constant enumeration} \\
		\vspace{6pt}\hspace{.5cm} \specialcell{
			\textit{Answer ``n'' to any prompts in the interactive} \\
			\textit{recursive removal of \underline{``dir1'', ``dir2'', and 
					``dir3''}}.} \\
		\textbf{Complex task} \\
		\vspace{6pt}\hspace{.5cm}\specialcell{
			\textit{Recursively finds all files in a current folder} 
			\\
			\textit{\underline{excluding already compressed files and 
					compresses}} \\
			\textit{\underline{them with level 9}}.} \\
		\textbf{Intelligible/Non-grammatical description} \\
		\vspace{6pt}\hspace{.5cm}\specialcell{
			\textit{Find all regular files in the current directory tree} \\
			\textit{\underline{and print a command to move them to the 
					current}} \\ 
			\textit{\underline{directory}}.}
	\end{tabular}
	\caption{Samples of natural language descriptions for the 
		major error causes.}\label{tb:error-class-examples}
\end{table}

\begin{table*}[!htbp]
	\small
	\centering
	\setlength\tabcolsep{2.5pt}
	\begin{tabular}{m{3.6cm}|m{11.4cm}|c}
		\hline
		Natural Language & Model Output & Model  \\
		\hline
		\multirow{7}{*}{\specialcell{\textit{Delete all empty directories} \\ 
				\textit{in minimum 2 levels down} \\
				\textit{the root directory}}} & \<find root -mindepth 2 -type d 
		-empty -delete> & Human \\
		\cline{2-3}
		& \<find root -mindepth 2 -type d -empty -delete> & C-Seq2Seq \\
		& \<find root -mindepth 2 -type d -empty -delete> & 
		C-CopyNet \\
		& \<find root -mindepth 2 -type d -empty -delete> & T-Seq2Seq \\
		& \<find root -mindepth 2 -type d -empty -delete> & T-CopyNet \\
		& \<find root -mindepth 2 -type d -empty -delete> & ST-Seq2Seq \\
		& \<find root -mindepth 2 -type d -empty -delete> & ST-CopyNet \\
		& \<find \red{\underline{.}} -mindepth 2 -type d -empty -delete> & 
		Tellina \\
		\hline
		\multirow{7}{*}{\specialcell{\textit{Find all Subscription.java} \\
				\textit{files/directories under} \\ 
				\textit{current directory and enter} \\
				\textit{into the parent directory of} \\
				\textit{the first one found}}} & \<cd \$(find . -name 
		Subscription.java 
		-printf '\%h\textbackslash n')> & Human \\
		\cline{2-3}
		& \<find . -name Subscription.java \incomplete> 
		& 
		C-Seq2Seq \\
		& \<find . -name Subscription.java | xargs \red{\underline{grep 
				Subscription.java}}> & C-CopyNet \\
		& \<cd \$(find . -name Subscription.java | xargs -I \ttcbs\ dirname 
		\ttcbs)> & 
		T-Seq2Seq \\
		& \<cd \$(find . -name Subscription.java | xargs -I \ttcbs\ dirname 
		\ttcbs)> & 
		T-CopyNet \\
		& \<cd \$(find . -name Subscription.java | xargs -I \ttcbs\ dirname 
		\ttcbs)> & 
		ST-Seq2Seq \\
		& \<cd \$(find . -name Subscription.java | xargs -I \ttcbs\ dirname 
		\ttcbs)> & 
		ST-CopyNet \\
		& \<cd \$(find . -name "Subscription.java" | xargs -I \ttcbs\ dirname 
		\ttcbs)> & 
		Tellina \\
		\hline
		\multirow{7}{*}{\specialcell{\textit{Add executable permission} \\
				\textit{to "pretty-print"}}} & \<chmod +x pretty-print> & Human 
		\\
		\cline{2-3}
		& \<chmod +x \red{\underline{prety-printf}}> & C-Seq2Seq \\
		& \<chmod \red{\underline{pretty-print pretty-print}}> & C-CopyNet \\
		& \<chmod +x \red{\underline{UNK}}> & T-Seq2Seq \\
		& \<chmod +x \red{\underline{UNK}}> & T-CopyNet \\
		& \<chmod +x \red{\underline{UNK-UNK-UNK}}> & ST-Seq2Seq \\
		& \<chmod +x pretty-print> & ST-CopyNet \\
		& \<chmod \red{\underline{[permission]}} "pretty-print"> & Tellina \\
		\hline
		\multirow{7}{*}{\specialcell{\textit{Search for the string 'git' in} \\ 
				\textit{all the files under current} \\ 
				\textit{directory tree without} \\
				\textit{traversing into '.git' folder} \\ 
				\textit{and excluding files that have} \\
				\textit{'git' in their names}}} & \<find . -path ./.git -prune 
		-o -not -name '*git*' -print |xargs grep git> & Human \\
		\cline{2-3}
		& \<find . -name \wrong{git} -prune -or -print \incomplete> 
		& 
		C-Seq2Seq \\
		& \<find . -name \*.git -prune -or \red{\underline{-name "*.git"}} 
		-print \incomplete> & C-CopyNet \\
		& \<find . -not -name "*.*" -not -path UNK \red{\underline{-not -path 
				UNK}} \incomplete> & T-Seq2Seq \\
		& \<find . \wrong{-not \textbackslash( -path UNK -prune 
				\textbackslash)} -or \textbackslash( -type f -and ! -name 
				\wrong{UNK} \textbackslash) -print0 | 
		xargs -0 -I \ttcbs\ grep \incomplete> & T-CopyNet \\
		& \<find . \red{\underline{-name '*.git'}} -not -path '.git' | xargs -I 
		\ttcbs\ grep 'git' \ttcbs> & ST-Seq2Seq \\
		& \<find . -not -name ".git" -not -path \red{\underline{"*.git*"}} -not 
		-name "*git*" | xargs -I \ttcbs\ grep git \ttcbs> & ST-CopyNet \\
		& \<find \red{\underline{"git"}} -not -path \red{\underline{".git"}} 
		-not -name \red{\underline{"*"}} | \red{\underline{grep "git"}}> & 
		Tellina \\
		\hline
		\multirow{7}{*}{\specialcell{\textit{Rename "file.txt" in} \\
				\textit{directories "v\_1", "v\_2"}, \\
				\textit{and "v\_3" each to "v\_1.txt"}, \\
				\textit{"v\_2.txt", and "v\_3.txt"} \\
				\textit{respectively and print the} \\
				\textit{conversion}}} & \<ls -d v\_{1,2,3} | xargs -i mv -v 
				\ttcbs/file.txt \ttcbs/\ttcbs.txt> & Human \\
		\cline{2-3}
		& \<\red{\underline{mv file.txt v\_1.txt}}> & C-Seq2Seq \\
		& \<\red{\underline{mv file.txt v\_3.txt}}> & C-CopyNet \\
		& \<\red{\underline{mv UNK UNK}}> & T-Seq2Seq \\
		& \<\red{\underline{mv UNK UNK}}> & T-CopyNet \\
		& \<\red{\underline{diff current\_1 \ttcbs}}> & ST-Seq2Seq \\
		& \<\red{\underline{ssh -i v\_1.txt v\_3.txt}}> & ST-CopyNet \\
		& \<\red{no output}> & Tellina \\
		\hline
	\end{tabular}
	\caption{Example predictions of the baseline 
		approaches. The prediction errors are underlined.}\label{tb:sample-output}
\end{table*}

\paragraph{Sparsity in Training Data} For both models, the top-one error cause 
is when the NL description maps to utilities or flags that rarely appeared in the train set 
(Table~\ref{tb:error-class-examples}). As mentioned in 
section~\ref{sec:domain}, the bash domain consists of a large number of 
utilities and flags and it is expensive to gather enough training data for 
all of them.

\paragraph{Common Errors of RNN Translation Models} The second major error 
class is commonly-known errors for RNN-based translation 
models (utility error, flag error and argument formatting error in 
Figure~\ref{fig:error-class}). When the RNN
mis-interprets or overlooks certain chunks of NL descriptions, 
the decoder can generate a wrong utility/flag or  
omit a utility/flag from the output sequence. Since the ST-CopyNet 
model also relies on the RNNs to generate sub-token contents, it suffers 
more from such problems --- the sub-token based models in general have more command structure 
errors and they frequently generated arguments that are a few edit distance 
away from the correct ones.
Interestingly, we noticed that few command template errors are syntax errors. The output commands often remain executable despite the semantic errors in different Bash components.

\paragraph{Constant Enumeration} In some cases, the NL descriptions contain
sequences of constant values as an enumeration of system objects or string patterns 
(Table~\ref{tb:error-class-examples}). We observed that both models struggled 
to extract all the constants correctly from this type of descriptions and usually failed
to set the extracted constants into the correct command slots. Moreover, long 
sequences of OOVs also down-qualify the RNN encodings and both models made more 
command structure errors in such cases.

\paragraph{Complex Task} We found several cases where the NL description
specifies a complex task and would be better broken into separate sentences 
(Table~\ref{tb:error-class-examples}). When the task gets complicated, the NL description 
gets verbose. As noted in 
previous work~\cite{dblp:journals/corr/bahdanaucb14}, the performance of RNNs decreases for longer sequences. 
Giving high-quality NL description for complex tasks are also more difficult for the users in practice --- multi-turn interaction is probably necessary for these cases.

\paragraph{Other Classes} For the rest of the error cases, we observed that the  
model failed to translate the specifications in \<()>, long descriptions of regular 
expressions and intelligible/non-grammatical NL descriptions 
(Table~\ref{tb:error-class-examples}).
There are also errors propogated from the pre-processing tools such as the NL 
tokenizer. In addition, the stage-wise system Tellina made a 
significant number of mistakes specific to its non-end-to-end modeling approach, e.g.
the limited coverage of its set of manually defined heuristic rules.

\begin{table*}[!htbp]
	\small
	\centering
	\setlength\tabcolsep{1.5pt}
	\begin{tabu}{c|c|ccccc|c|c|c|c}
		\hline
		\multirow{2}{*}{Dataset} & \multirow{2}{*}{PL} & \# & \# & \# & Avg. \#
		& Avg. \# & NL & Code & Semantic & Introduced \\
		& & pairs & words & tokens & w. in nl & t. in code & collection & 
		collection & alignment & by \\
		\hline
		{IFTTT} & {DSL} & 86,960 & -- & -- & 7.0 & 21.8 & 
		\multirow{4}{*}{scraped} & 
		\multirow{4}{*}{scraped} & 
		\multirow{3}{*}{Noisy} & {\cite{dblp:conf/acl/quirkmg15}} \\
		\cline{11-11}
		
		{C\#2NL*} & {C\#} & 66,015 & 24,857 & 91,156 & 12 & 38 & & & &
		\multirow{2}{*}{{\cite{DBLP:conf/acl/IyerKCZ16}}} \\
		
		{SQL2NL*} & {SQL} & 32,337 & 10,086 & 1,287 & 9 & 46 & & & & \\
		\cline{10-11}
		
		RegexLib & Regex & 3,619 & 13,491 & 179$^\text{\ding{89}}$ & 36.4 & 
		58.8$^\text{\ding{89}}$ & & & \multirow{3}{*}{Good$^\text{\ding{64}}$} 
		& \cite{Zhong2018GeneratingRE} \\
		\cline{1-9}\cline{11-11}
		
		{HeartStone} & Python & 665 & -- & -- & 7 & 352$^\text{\ding{89}}$ & 
		\multirow{2}{*}{\specialcell{game card \\ description}} & 
		\multirow{2}{*}{\specialcell{game card \\ source code}} & &
		\multirow{2}{*}{{\cite{DBLP:conf/acl/LingBGHKWS16}}} \\
		
		{MTG} & Java & 13,297 & -- & -- & 21 & 1,080$^\text{\ding{89}}$ & & & 
		& \\ 
		\cline{11-11}

		\tabucline[1.5pt]{-}
		
		\multirow{2}{*}{StaQC} & Python & 147,546 & 17,635 & 137,123 & 9 & 86 & 
		extracted & extracted & \multirow{2}{*}{Noisy} & 
		\multirow{2}{*}{\cite{StaQC}}\\
		& SQL & 119,519 & 9,920 & 21,413 & 9 & 60 & using ML & using ML & & 
		\\
		\tabucline[1.5pt]{-}
		
		{NL2RX} & {Regex} & 10,000 & 560 & 
		45$^\text{\ding{89}}$$^\text{\ding{61}}$ & 10.6 & 
		26$^\text{\ding{89}}$ & synthesized \& & \multirow{2}{*}{synthesized} 
		& \multirow{2}{*}{\specialcell{Very \\ Good}} & 
		{\cite{DBLP:conf/emnlp/LocascioNDKB16}} \\
		\cline{11-11}
		
		{WikiSQL} & {SQL} & 80,654 & -- & -- & -- & -- & paraphrased & & &
		{\cite{zhong2017seq2sql}} \\
		\tabucline[1.5pt]{-}
		
		\multirow{2}{*}{NLMAPS} & \multirow{2}{*}{DSL} & \multirow{2}{*}{2,380} & \multirow{2}{*}{1,014} & 
		\multirow{2}{*}{--} & \multirow{2}{*}{10.9} & \multirow{2}{*}{16.0} & synthesized & expert & 
		\multirow{10}{*}{\specialcell{Very \\ Good}} & 
		\multirow{2}{*}{\cite{DBLP:conf/naacl/HaasR16}} \\
		& & & & & & & given code & written  & \\
		\cline{1-9}\cline{11-11}
		
		{Jobs640$^\text{\ding{72}}$} & {DSL} & 640 & 391 & 58$^\text{\ding{61}}$ & 9.8 & 22.9 & 
		\multirow{4}{*}{user written}& \multirow{6}{*}{
		\specialcell{expert \\ written \\ given NL}} & & {\cite{DBLP:conf/ecml/TangM01}} \\
		\cline{11-11} 
		
		{GEO880} & {DSL} & 880 & 284 & 60$^\text{\ding{61}}$ & 7.6 & 19.1 & & & 
		& {\cite{Zelle:1996:LPD:1864519.1864543}} \\
		\cline{11-11}
		
		{Freebase917} & {DSL} & 917 & -- & -- & -- & -- & & & & {\cite{DBLP:conf/acl/CaiY13}} \\
		\cline{11-11}
		
		{ATIS$^\text{\ding{72}}$} & {DSL} & 5,410 & 936 & 176$^\text{\ding{61}}$ & 11.1 & 28.1 & & 
		& & {\cite{dahl:1994:esa:1075812.1075823}} \\
		\cline{1-8}\cline{11-11}
		
		WebQSP & DSL & 4,737 & -- & -- & -- & -- & search log & & & 
		\cite{DBLP:conf/acl/YihRMCS16} \\
		\cline{1-8}\cline{11-11}
		
		{NL2RX-KB13} & {Regex} & 824 & 715 & 
		85$^\text{\ding{89}}$$^\text{\ding{61}}$ & 7.1 & 
		19.0$^\text{\ding{89}}$ & turker written & & & 
		{\cite{dblp:conf/naacl/kushmanb13}} \\
		\cline{1-9}\cline{11-11}
		Django$^\text{\ding{75}}$ & Python & 18,805 & -- & -- & 14.3 & -- &
		expert written & \multirow{2}{*}{scraped} & & 
		\cite{DBLP:conf/kbse/OdaFNHSTN15} \\
		\cline{1-7}\cline{11-11} 
		
		NL2Bash  & Bash & 9,305 & 7,790 & 6,234 & 11.7 & 7.7 & given code& & & 
		Ours\\
		\hline
		
	\end{tabu}
	\caption{Comparison of datasets for translation of natural language to (short) code snippets.
	\small{*: Both C\#2NL and SQL2NL were originally collected to train systems that explain
	code in natural language. $^\text{\ding{89}}$: The 
	code length is counted by characters instead of by tokens. 
	$^\text{\ding{61}}$: When 
	calculating \# tokens for these datasets, the open-vocabulary constants 
	were replaced with positional placeholders. $^\text{\ding{64}}$: These 
	datasets were collected from sources where the NL and code exist 
	pairwise, but the pairs were not 
	compiled for the purpose of semantic parsing. $^\text{\ding{72}}$: Both 
	Jobs640 and ATIS consist of mixed manually-generated and 
	automatically-generated NL-code pairs. $^\text{\ding{75}}$ The Django 
	dataset consists of pseudo-code/code pairs.}}\label{tb:comparison}
\end{table*}

Based on the error analysis, we recommend future work 
to build shallow command structures in the decoder instead of synthesizing the entire output in sequential manner, e.g. using separate RNNs for template translation and argument filling. 
The training data sparsity can possibly be alleviated by 
semi-supervised learning using unlabeled Bash commands
or external resources such as the Linux man pages. 

\section{Comparison to Existing Datasets}
\label{sec:comparison}

This section compares NL2Bash to other commonly-used semantic parsing and 
NL-to-code datasets.\footnote{We focus on generating utility commands/scripts 
from natural language and omitted the datasets in the domain of programming 
challenges~\cite{2018arXiv180204335P} and code base 
modeling~\cite{Nie2018NaturalLP}.
        }
We compare the datasets with respect to: 
(1) the programming language used, (2) size, (3) shallow quantifiers of 
difficulty (i.e. \# unique NL words, \# unique program tokens, 
average length of text and average length of code) and (4) 
collection methodology.
Table~\ref{tb:comparison} summarizes the 
comparison. We directly quoted the published dataset statistics we have found, 
and computed the statistics of other released datasets to our best effort. 

\paragraph{Programming Languages} Most 
of the datasets were constructed for domain-specific languages 
(DSLs). Some of the recently proposed datasets use Java, Python, C\#, and Bash, 
which are Turing-complete programming languages. This shows the beginning of an 
effort to apply natural language 
based code synthesis to more general PLs.

\paragraph{Collection Methodology}
Table~\ref{tb:comparison} sorts the datasets by
increasing amount of manual effort spent on the data collection. NL2Bash is by 
far the largest dataset constructed using practical code snippets and 
expert-written natural language. In addition, it 
is significantly more diverse (7,790 unique words and 6,234 unique command 
tokens) 
compared to other manually constructed datasets. 

The approaches of automatically scraping/extracting parallel 
natural language and code have been adopted more
recently. A major resource of such parallel data are
question-answering 
forums (StackOverflow: \url{https://stackoverflow.com/}) and 
cheatsheet 
websites (IFTTT: \url{https://ifttt.com/} and 
RegexLib: \url{http://www.regexlib.com/}). Users post code snippets 
together with 
natural language questions or descriptions in these venues. The problem with 
these data is that they are loosely aligned and cannot be directly used for 
training. Extracting good alignments from them is very 
challenging~\cite{dblp:conf/acl/quirkmg15,DBLP:conf/acl/IyerKCZ16,StaQC}. That 
being said, these datasets significantly surpasses 
the manually gathered ones in terms of size and diversity, hence demonstrating 
significant potential for 
future work. 

Alternatively, Locascio et al. (2016) and  Zhong et al. (2017a)
proposed synthesizing parallel natural language and 
code using a synchronous grammar. They also hired Amazon Mechanical Turkers to 
paraphrase 
the synthesized natural language sentences in order to increase their 
naturalness and 
diversity. While 
the synthesized domain may be less diverse compared to naturally existed 
ones, they served as an excellent resource for data augmentation or zero-shot 
learning. The downside is that developing synchronous grammars for domains 
other than simple DSLs is challenging, and other data collection methods 
are still necessary for them.

The different data collection methods are complimentary and we expect to see 
more future work mixing different strategies.

\section{Conclusions}

We studied the problem of mapping English sentences to Bash commands (NL2Bash), 
by introducing a large new dataset and baseline methods. 
NL2Bash is by far the largest NL-to-code dataset constructed using 
practical code snippets and expert-written natural language.
Experiments demonstrated 
competitive performance of 
existing models as well as significant room for future 
work on this challenging semantic parsing problem. 

\section{Acknowledgements}
The research was supported in part by DARPA under the DEFT program 
(FA8750-13-2-0019), the ARO (W911NF-16-1-0121), the NSF (IIS1252835, 
IIS-1562364), gifts from Google and Tencent, and an Allen Distinguished 
Investigator Award. 
We thank the freelancers who worked with us to make the corpus.
We thank Zexuan Zhong for providing us the statistics of the RegexLib 
dataset.
We thank the anonymous reviewers, Kenton Lee, Luheng He, Omer Levy for 
constructive feedback on the paper draft, and the UW NLP/PLSE groups 
for helpful conversations.


\section{Bibliographical References}
\label{main:ref}
\bibliographystyle{lrec}
\bibliography{main}

\newpage
\section*{Appendices}
\label{sec:appendix}
\addcontentsline{toc}{section}{Appendices}
\renewcommand{\thesubsection}{\Alph{subsection}}
\subsection{Additional Data Statistics}
\label{subsec:data-stats-cont}

\subsubsection{Distribution of Less Frequent Utilities}
\begin{figure}[ht]
	\centering
	\includegraphics[width=0.9\linewidth]{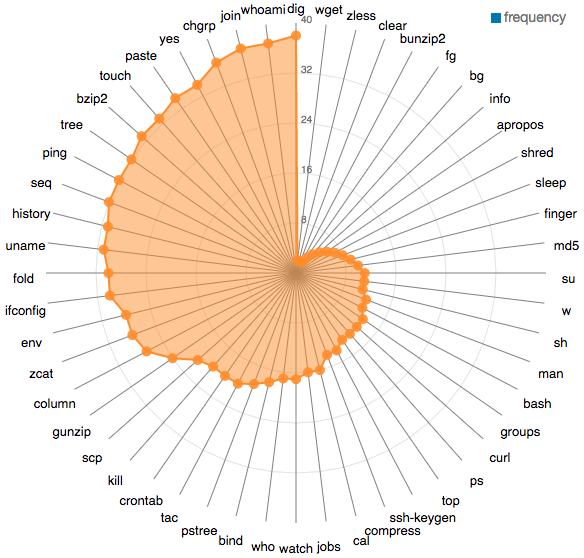}
	\caption{Frequency radar chart of the 52 least frequent bash utilities in 
	the datasets.}
	\label{fig:utility-freq-low}
	\eat{\todo{It's weird that this diagram is circular but
			\cref{fig:utility-flag-number} represents the same data as a 
			histogram.  I
			would make the two consistent (use a histogram here).}
		\todo{Ensure that the actual number can be determined from this diagram,
			just as it can for \cref{fig:utility-flag-number}.  (The units 
			aren't
			even labeled on this.)}}
\end{figure}

Figure~\ref{fig:utility-freq-low} illustrates the frequencies of the 52 least 
frequent bash utilities in our dataset. Among them, the most frequent utility 
\<dig> appeared only 38 times in the dataset. 7 utilities appeared 5 times or 
less. We discuss in the next session that many of such low frequent utilities 
cannot be properly learned at this stage, since the limited number of training 
examples we have cannot cover all of their usages, or even a reasonably 
representative subset.

\subsubsection{Flag Coverage}
Table~\ref{tb:flag-coverage} shows the total number of flags (both 
long and short) a utility has and the number of flags of the utility that appeared in the training set. 
We show the statistics for the 10 most and least frequent utilities in the corpus. 
We estimate the total number of flags a utility has by the number of flags we
manually extracted from its GNU man page. The estimation is a lower bound as we might miss certain flags due to man page version mismatch and human errors.

\begin{table}[ht]
	\centering
	\begin{tabular}{cccc}
		\multirow{2}{*}{Utility} & \multirow{2}{*}{\# flags} & \# flags \\
		& & in train set \\
		\hline
		find & 103 & 68 \\
		xargs & 32 & 15 \\
		grep & 82 & 42 \\
		rm & 17 & 7 \\
		echo & 5 & 2 \\
		sort & 50 & 19 \\
		chmod & 14 & 4 \\
		wc & 13 & 6 \\
		cat & 19 & 4 \\
		\hline
		sleep & 2 & 0 \\
		shred & 17 & 4 \\
		apropos & 30 & 0 \\
		info & 34 & 2 \\
		bg & 0 & 0 \\
		fg & 0 & 0 \\
		wget & 171 & 2 \\
		zless & 0 & 0 \\
		bunzip2 & 14 & 0 \\
		clear & 0 & 0 \\
	\end{tabular}
	\caption{Training set flag coverage. The upper-half of the table shows the 
	10 most frequent utilities in the corpus. The lower-half of the table shows 
	the 10 least frequent utilities in the corpus.}\label{tb:flag-coverage}
\end{table}

Noticed that for most of the utilities, only less than 
half of their flags appear in the train set. One reason contributed to the small coverage is that most command flags has a full-word replacement for readability (e.g. the 
readable replacement for \<-t> of \<cp> is \<--target-directory>), yet most 
Bash commands written in practice uses the short flags. We could solve this 
type of coverage problem by normalizing the commands to contain only 
the short flags. (Later we can use deterministic rules to show the readable version to the user.) 
Nevertheless, for many utilities a subset of their flags are 
still missing from the corpus. Conducting zero-shot learning for those missing 
flags is an interesting future work.

\subsection{Data Quality}
\label{subsec:data-quality}
We asked two freelancers to evaluate 100 text-command pairs sampled from our 
train set. The freelancers did not author the sampled set of pairs 
themselves. We asked 
the freelancers to judge the correctness of each pair. We also asked the 
freelancers to judge if the natural language description is clear enough for 
them to understand the descriptor's goal. We then manually examined the 
judgments made by the two freelancers and summarize the findings below.

\begin{table*}[t]
	\centering
	\begin{tabular}{l}
		\hline
		\textit{Find all \wrong{executables} under /path directory} \\
		\vspace{6pt}\hspace{.5cm} \<find /path -perm /ugo+x> \\
		\vspace{6pt}\hspace{.5cm} \specialcell{
			``Executables generaly means executable 
			files, thus needs \<-type f>. Also, \</ugo+x> should be 
			\<-ugo+x>. The current \\ 
			command lists all the directories too as directories generally have 
			execute permission at least for the owner \\ (\</ugo+x> allows 
			that, while \<-ugo+x> would require execute permission for all).''
		} \\
		\textit{Search the current directory tree for all regular 
			\wrong{non-hidden files} except *.o} \\
		\vspace{6pt}\hspace{.5cm} \<find ./ -type f -name "*" -not -name "*.o"> 
		\\
		\vspace{6pt}\hspace{.5cm} \specialcell{
			``Criteria not met: non-hidden, requires something like  -not -name 
			\<'.*'>.''
		} \\
		\textit{Display \wrong{all the text files from the current folder and 
				skip searching in skipdir1 and skipdir2 folders}} \\
		\vspace{6pt}\hspace{.5cm} \<find . $\backslash$( -name skipdir1 -prune 
		, -name skipdir2 -prune -o -name "*.txt" $\backslash$) -print> \\
		\vspace{6pt}\hspace{.5cm} \specialcell{
			``Result includes \<skipdir2> (this directory name only), the  
			\<-o> can be replaced with comma , to solve this.''
		} \\
		\textit{Find all the files that have been modified in the last 2 days 
			\wrong{(missing \<-daystart> description)}} \\ 
		\vspace{6pt}\hspace{.5cm} \<find . -type f -daystart -mtime -2> \\
		\vspace{6pt}\hspace{.5cm} \specialcell{
			``\<daystart> is not specified in description.''
		} \\
		\textit{Find all the files that have been modified \wrong{since the 
				last time we checked}} \\ 
		\vspace{6pt}\hspace{.5cm} \<find /etc -newer /var/log/backup.timestamp 
		-print> \\
		\vspace{6pt}\hspace{.5cm} \specialcell{
			```Since the last time we checked', the backup file needs to be 
			updated after the command completes to make this \\
			possible.''
		} \\
		\textit{Search for all the .o files in the current directory which have 
			permisssions 664 and print them.} \\ 
		\vspace{6pt}\hspace{.5cm} \<find . -name *.o -perm 664 -print> \\
		\vspace{6pt}\hspace{.5cm} \specialcell{
			``Non-syntactical command. Should be \<\*.o> or \<"*.o">.''
		} \\
		\textit{Search for text files in the directory "/home/user1" and 
			\wrong{copy 
				them to the directory /home/backup}} \\ 
		\vspace{6pt}\hspace{.5cm} \<find /home/user1 -name '*.txt' | xargs cp 
		-av --target-directory=/home/backup/ --parents> \\
		\vspace{6pt}\hspace{.5cm} \specialcell{
			``\<--parents> not specified in description, it creates all the 
			parent dirs of the files inside target dir, e.g, a file named \\
			\<a.txt> would be copied to \</home/backup/home/user1/a.txt>.''
		} \\
		\specialcell{\textit{Search for the regulars file starting with HSTD 
				\wrong{(missing case insensitive description)} which have been 
				modified} \\
			\textit{yesterday from day start and copy them to /path/tonew/dir}} \\ 
		\vspace{6pt}\hspace{.5cm} \<find . -type f -iname 'HSTD*' -daystart 
		-mtime 1 -exec cp \ttcbs /path/to new/dir/ $\backslash$;> \\
		\hspace{.5cm} \specialcell{
			``Case insensitive not specified but \<-iname> used, extra spaces 
			in \</path/to new/dir/>.''
		} \\
		\hline
	\end{tabular}
	\caption{Training examples whose NL description has errors (underlined). The error explanation is written by the freelancer.}\label{tb:train-error}
\end{table*}

The freelancers identified errors in 15 of the sampled training pairs, which 
results in approximately 85\% annotation accuracy of the training data. 3 of 
the errors are caused by the fact that some utilities (e.g. \<rm>, \<cp>, 
\<gunzip>) handle directories differently from regular files, but the natural 
language description failed to clearly specify if the target objects include directories or not. 4 cases were typos made by our annotators 
when copying
the constant values in a command to their descriptions. Being able to 
automatically detect constant mismatch may reduce the number of such errors. 
(Automatic mismatch detection can be directly added to the annotation interface.)
The rest of the 8 cases were caused by the annotators mis-interpreted/omitted 
the function of certain flags/reserved tokens or failed to spot syntactic 
errors in 
the command (listed in Table~\ref{tb:train-error}). For many of these cases, 
the Bash commands are only of medium 
length --- this shows that accurately describing all 
the information in a Bash command is still an error-prone task for Bash 
programmers. Moreover, some 
annotation mistakes are more thought-provoking as the operations in those examples might be 
difficult/unnatural for the users to describe at test time.
In these cases we should solicit the necessary information from the users through alternative ways, e.g. 
asking multi-choice questions for specific options or asking the user for examples.

Only 1 description was marked as ``unclear'' by one of the freelancers. The 
other freelancer still judged it as ``clear''. Similar trend were observed 
during the manual evaluation --- the freelancers have little problem understanding each 
other's descriptions.

It is worth noting that while we found 15 wrong pairs out of 100, for 13 of them the annotator only misinterpreted one of the
command tokens. Hence the overall performance of the annotators is high, especially given the large domain size.


\subsection{Automatic Evaluation Results}
\label{subsec:auto-eval}

\begin{table}[t]
	\centering
	\setlength\tabcolsep{4.5pt}
	\begin{tabular}{c|c|cc|cc}
		\hline
		\multicolumn{2}{c|}{Model} &\BLEU{1} &\BLEU{3} & \TM{1} &\TM{3}\\
		\hline
		\multirow{3}{*}{Seq2Seq} 
		& Char 	& 49.1	& 56.7 & 0.57 	& 0.64 	\\
		& Token	& 36.1	& 43.9 & 0.65	&\highest{0.75} \\
		& Sub-token & 46	& 52	& 0.65	& 0.71 	\\
		\hline
		\multirow{3}{*}{CopyNet} 
		& Char 	& 49.1	& 56.8 & 0.54 	& 0.61 	\\
		& Token & 44.9	& 54.2	& 0.65	&0.74	\\
		& Sub-token & \highest{55.3}	& \highest{61.8} & 0.64	& 0.71	\\
		\hline
		\multicolumn{2}{c|}{Tellina} & 46	& 52 &0.61  &0.70\\
		\hline
	\end{tabular}
	\caption{Automatically measured performance of the baseline systems on the 
	full dev set.}
	\label{tb:baselines-dev-auto}
\end{table}

We report two types of fuzzy evaluation metrics automatically computed over full 
dev set in table~\ref{tb:baselines-dev-auto}. We define TM as the 
maximum percentage of close-vocabulary token (utilities, flags and reserved 
tokens) overlap between a predicted command and the reference commands. 
(TM is a command structure accuracy measurement.) \TM{k} is the maximum TM 
score achieved by the top-{$k$} candidates generated by a system. We use BLEU 
as an approximate measurement for full command accuracy. \BLEU{k} is the 
maximum BLEU score achieved by the top-{$k$} candidates generated by a system.

First, we observed from table~\ref{tb:baselines-dev-auto} that while the 
automatic evaluation metrics agrees with the manual ones
(Table~\ref{tb:baselines}) on the system with the highest full command 
accuracy and the system with the highest command structure accuracy, they do 
not agree with the manual evaluation in all cases (e.g. character-based models 
have the second-best BLEU score). Second, the TM score is not discriminative 
enough -- several systems scored similarly on this metrics.


\end{document}